\documentclass[10pt,twocolumn,letterpaper]{article}

\usepackage{cvpr}              

\usepackage[dvipsnames]{xcolor}

\usepackage{algorithm}
\usepackage{algpseudocode}

\usepackage{amsmath,amsfonts,bm}

\def\eqref#1{equation~\ref{#1}}

\def\1{\bm{1}}

\DeclareMathAlphabet{\mathsfit}{\encodingdefault}{\sfdefault}{m}{sl}
\SetMathAlphabet{\mathsfit}{bold}{\encodingdefault}{\sfdefault}{bx}{n}

\newcommand{\R}{\mathbb{R}}

\usepackage{makecell}

\definecolor{cvprblue}{rgb}{0.21,0.49,0.74}
\usepackage[pagebackref,breaklinks,colorlinks,citecolor=cvprblue]{hyperref}

\def\paperID{80} 
\def\confName{CVPR}
\def\confYear{2024}

\title{CUE-Net: Violence Detection Video Analytics with Spatial Cropping, Enhanced UniformerV2 and Modified Efficient Additive Attention}

\author{Damith Chamalke Senadeera$^{1,2}$, Xiaoyun Yang$^{3}$, Dimitrios Kollias$^{1,2}$, Gregory Slabaugh$^{1,2}$\\
$^{1}$School of Electronic Engineering and Computer Science, Queen Mary University of London, UK\\
$^{2}$Queen Mary's Digital Environment Research Institute (DERI), London, UK\\
$^{3}$Remark AI UK Limited, London, UK\\
{\tt\small d.c.senadeera@qmul.ac.uk, xiaoyun.yang@remarkai.co.uk, \{d.kollias, g.slabaugh\}@qmul.ac.uk}
}

\begin{document}
\maketitle
\begin{abstract}
In this paper we introduce CUE-Net, a novel architecture designed for automated violence detection in video surveillance. As surveillance systems become more prevalent due to technological advances and decreasing costs, the challenge of efficiently monitoring vast amounts of video data has intensified. CUE-Net addresses this challenge by combining spatial \textbf{C}ropping with an enhanced version of the \textbf{U}niformerV2 architecture, integrating convolutional and self-attention mechanisms alongside a novel Modified \textbf{E}fficient Additive Attention mechanism (which reduces the quadratic time complexity of self-attention) to effectively and efficiently identify violent activities. This approach aims to overcome traditional challenges such as capturing distant or partially obscured subjects within video frames. By focusing on both local and global spatio-temporal features, CUE-Net achieves state-of-the-art performance on the RWF-2000 and RLVS datasets, surpassing existing methods. The source code is available at \footnote[1]{https://github.com/damith92/CUENet}.
\end{abstract}    
\section{Introduction}
\label{sec:intro}

According to the World Bank, there has been an increase in the worldwide crime rate in the last five years \cite{MacroTrends}. Surveillance cameras are often deployed to help deter violence, provide real-time monitoring and collect evidence of criminal or violent activity.  Thanks to advances in technology, surveillance systems are becoming increasingly affordable and easier to deploy. As the number of these deployed surveillance cameras grows, it rapidly becomes expensive and challenging for human operators to manually monitor camera feeds \cite{review_1, review_2}. Therefore there is substantial need for automated approaches to monitor surveillance cameras, simplifying the process of Violence Detection (VD) in a more accurate and an efficient manner \cite{review_1, review_3}.

\begin{figure*}
  \centering
   \includegraphics[width=0.8\linewidth]{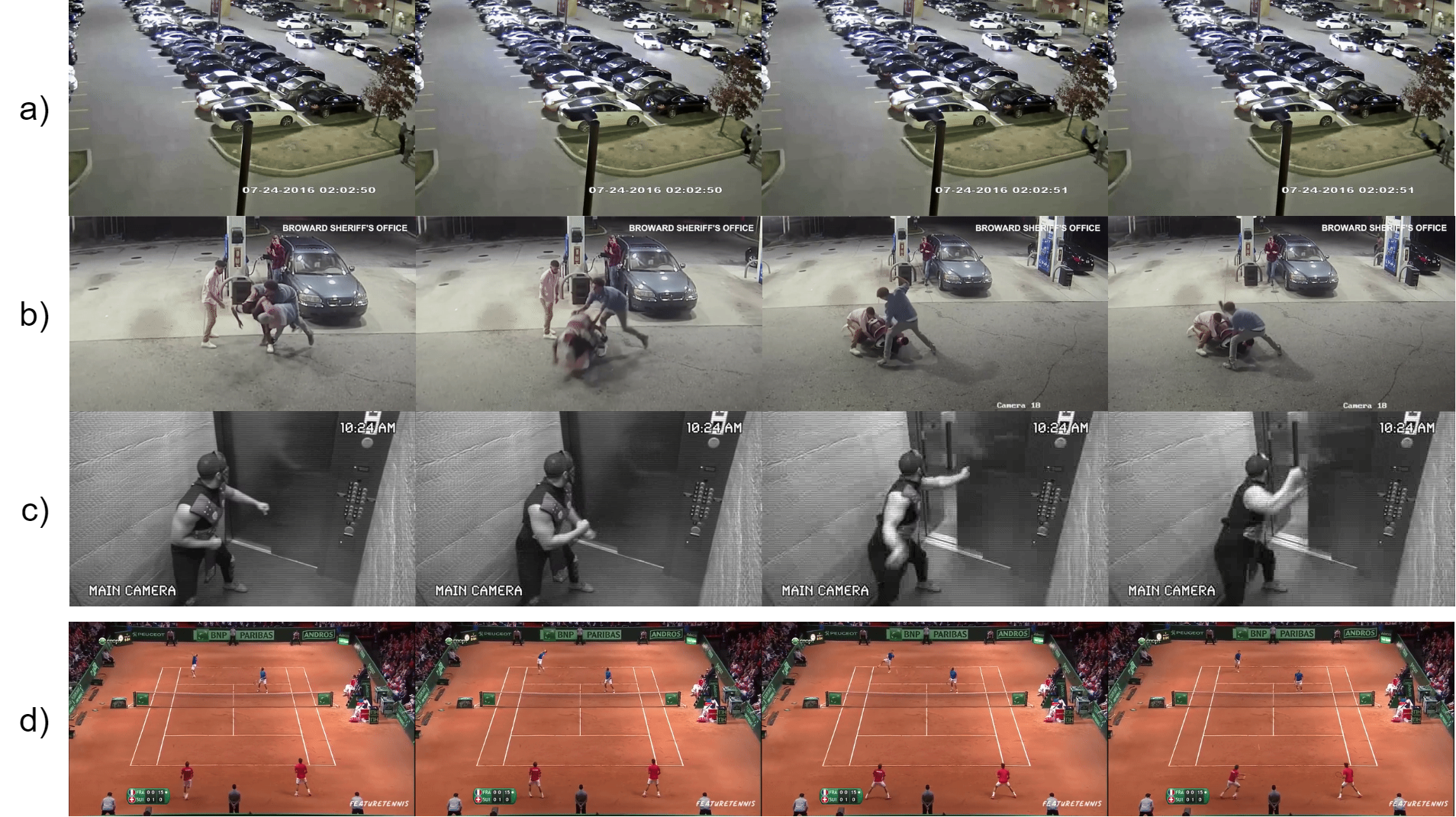}
   \caption{Sample violence detection videos. \textbf{(a)} is a set of frames from a challenging video from RWF-2000 where the people involved in the violent incident are far away from the camera, occupying only a small part of the frame. \textbf{(b)} shows a typical violent video from the RWF-2000 dataset correctly classified by CUE-Net. \textbf{(c)} is a video from the RWF-2000 dataset test split, where a man makes punching actions but is not really engaging in a fight. CUE-Net incorrectly classifies this as a violent video. \textbf{(d)} is a video from the RLVS dataset which CUE-Net correctly classifies as non-violent, but for which the ground-truth is mislabeled as violent.}
  \label{fig:7}
\end{figure*}

To respond to the challenge of efficient, automated violence detection from video, effective computer vision methods are required. Deep learning techniques such as Convolutional Neural Networks (CNNs) and more recently Transformer-based architectures have shown a great promise in solving computer vision related automated violence detection \cite{review_1, review_2, review_3}. The success of violence detection is highly dependent on the objects and people present in the captured videos~\cite{review_1, review_2}. Detection is difficult when the relevant features of the violent incidents are not captured properly, for example when the people involved in the violent  incident are far away and occupy only a small part of the frame, as seen in one of the example videos from the RWF-2000 dataset~\cite{rwf_dataset} in \cref{fig:3} (a). Although different mechanisms have been explored for automated violence detection, the opportunity for improvement remains due to challenges such as tracking and extracting fast moving people or objects involved in violence, low resolution scenarios and occlusion-related issues  \cite{review_1}.

Another research question relates to finding an effective and a robust processing architecture for violence detection in videos. An ideal architecture would be simultaneously capable of capturing the locally and globally important features across the temporal and the spatial dimensions. As discussed in \cite{action_recognition_VD_1, action_recognition_VD_2} CNN-based architectures have shown to better capture locally important features but not the globally important ones;  whereas \cite{swintransformer_VD} argues that the self-attention mechanism in the transformer architecture seems to better capture globally important features temporally. However,  transformer architectures may struggle with video data due to their quadratic computational complexity \cite{swiftformer}. Therefore, a novel solution which combines the advantages of convolutions to capture  local temporal features and transformers to capture global features using lightweight attention mechanisms is worthwhile exploring.

In this paper, we propose a novel architecture named CUE-Net which amalgamates spatial \textbf{C}ropping, with an enhanced version of the \textbf{U}niformerV2\cite{uniformerv2} architecture which incorporates the benefits of both the convolution and self-attention. In this architecture, we propose Modified \textbf{E}fficient Additive Attention (MEAA), a novel efficient attention mechanism which reduces the quadratic time complexity of self-attention to capture the important global spatio-temporal features, to mitigate the above mentioned bottlenecks. For the best of our knowledge, this is the first time that such a model which incorporates convolution and self-attention along with modified Efficient Additive Attention mechanism has been investigated in the context of violence detection in videos. Our contributions are as follows:

\begin{enumerate}

    \item We propose CUE-Net, a novel architecture for violence detection video analytics which incorporates a novel enhanced version of the UniformerV2 architecture along with Modified Efficient Additive Attention (MEAA), a novel attention mechanism to capture the important global spatio-temporal features.

    \item We incorporate a spatial cropping mechanism based on the detected number of people in our algorithm before the video is fed into the main learning algorithm, to focus the method on the area where violence is occurring without losing the important surrounding information.

    \item Our results set a new state-of-the-art on the RWF-2000 and RLVS datasets, outperforming the most recently published methods.

\end{enumerate}

\section{Related Work}
\label{sec:relatedwork}

This section summarizes the current state-of-the-art methods for VD and categorizes different methods used in the context of violence detection as an action recognition task vs an anomaly detection task.

\subsection{Deep Learning Architectures for Violence Detection using Anomaly Detection}

In anomaly detection scenarios, violent events are considered as scarce abnormal events deviating from normal day-to-day events. Algorithms learn to characterise the features of normal events, and violence detection is based on detecting events that do not lie in the normal distribution. However, in practice, the boundary between normal and anomalous behaviors can be ambiguous. Under realistic situations, similar behaviors may be normal or anomalous given different conditions, for example the the action of punching will be normal for a friendly fist bump but anomalous for a violent punch \cite{review_1, review_3}. The work of \cite{ucfcrime_dataset} proposes to learn anomalies through a deep Multiple Instance Learning (MIL) framework that treats a video as a bag with short segments/clips of each video as instances in a bag. However, \cite{anaomaly_VD_3} argues that the recognition of the anomalous instances is largely biased by the dominant normal (non-violent) instances of the data, especially when the abnormal events are subtle anomalies that exhibit only small differences compared with normal events. 


When trying to frame the VD problem in an anomaly detection context, violent (anomalous) events are identified by focusing mainly on learning how a normal situation looks like rather than focusing on the context of the violent behaviour.  Often violence is dependent on context as well as the actions happening in the scene. Models trained to detect anomalies in this manner might not adequately understand the context in which certain violent actions are taking place and therefore may not be able to generalize well as their main task is not to learn the context-specific features for violent events \cite{MIL_Survey}.

\begin{figure*}
  \centering
   \includegraphics[width=0.92\linewidth]{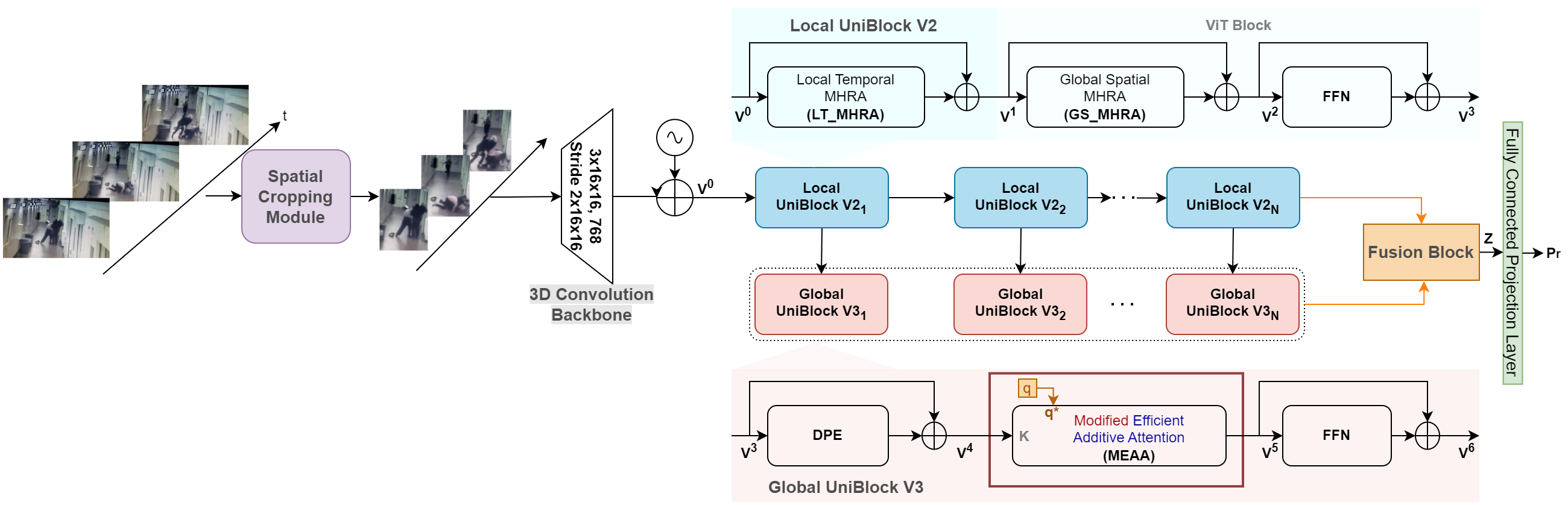}
   \caption{The overall CUE-Net architecture with its main components. \textbf{(a)} the Spatial Cropping Module uses the YOLO V8 algorithm to detect people and crop the video spatially; \textbf{(b)} the 3D Convolutional Block which is used to encode and downsample the frames spatio-temporally; \textbf{(c)} the Local UniBlock V2 which is mainly used to capture the important local dependencies with its main components LT\_MHRA, GS\_MHRA and a feed forward network (FFN); \textbf{(d)} the Global UniBlock V3 which is mainly used to capture the important global spatio-temporal dependencies, with its main components Dynamic Positional Embedding (DPE) unit, MEAA unit which implements a novel efficient self-attention mechanism and a feed forward network (FFN); \textbf{(e)} the Fusion Block which is used to fuse the outputs of the Local UniBlock V2 and Global UniBlock V3.}
  \label{fig:3}
\end{figure*}

\subsection{Deep Learning Architectures for Violence Detection in an Action Recognition Context}

Work by \cite{action_recognition_VD_1} introduced one of the earliest uses of 3D-CNNs along with a softmax classifier for violence detection. As a pre-processing step, first, frames where people are present are identified, with the premise that the violent actions will happen only when people are present. Then a 3D-CNN extracted spatio-temporal features out of the filtered frames and a soft-max layer classified the results. In another study, \cite{action_recognition_VD_3} introduced a novel approach for violence detection in the space of action recognition by learning contextual relationships between people using human skeleton points. Unlike the previous references, \cite{action_recognition_VD_3}  formulated 3D skeleton point clouds from human skeleton sequences extracted from videos and then performed interaction learning on these 3D skeleton point clouds, considering them as non-Euclidean graphs using Graph CNNs. \cite{action_recognition_VD_3} is one of the first papers to evaluate performance on a real-world surveillance violence detection data set (RWF-2000) \cite{rwf_dataset} where all most all the previous literature was evaluated on non-surveillance based datasets such as the Hockey Fight dataset \cite{hockeyfight}. \cite{action_recognition_VD_4} introduced a novel deep architecture comprising of two simultaneous pipelines, one to extract the skeletons of people using a pose estimation model and the other to estimate the dynamic temporal changes between frames where the outputs from the two pipelines were fused together using addition to transmit information even when one of the inputs provides a zero-valued signal. 

The current state-of-the-art approach for violence detection on the RWF-2000 dataset relies on a Video Swin Transformer \cite{swintransformer_VD}. This work applies a method to extract keyframes from the videos based on frame colour, texture and motion features using colour histograms, gray level co-occurrence matrices and optical flow. Then, a Video Swin Transformer \cite{videoswintrf} starts with processing small patches of the videos and gradually merges them into deeper transformer layers in spatio-temporal context, creating a hierarchical representation. This approach has enabled the aggregation of features from a local to a global context.

In summary, framing the violence detection problem using action recognition has advantages over anomaly detection. Recent literature has focused more on extracting rich and representative features of violent actions and derive a better contextual understanding in order to separate violent actions from normal activities \cite{review_1}.

\section{Proposed Method}
\label{sec:method}

In this section we first motivate our work and then discuss our proposed CUE-Net method in detail.

\textbf{Motivation}: Our work takes inspiration from the action recognition literature, as it provides an effective supervised method for video action recognition. In the action recognition space, a novel deep architecture called Unified Transformer (UniFormer) \cite{uniformerv1} has been introduced which seamlessly integrates the merits of 3D convolution and spatio-temporal self-attention in a concise format by implementing modules of both convolution and self-attention together to achieve a balance between computational complexity and accuracy. Later, the Uniformer Version 2 (UniformerV2) architecture~\cite{uniformerv2} modified these modules of the previous Uniformer architecture to implement them simultaneously and fuse at the end of the pipeline to capture the relevant spatio-temporal features. Also, UniformerV2 takes advantage of pre-trained ViT embeddings to initialize segments of the architecture to better make use of pretrained knowledge from large image datasets. 

However, self-attention has a quadratic computational complexity with respect to the sequence length, making it challenging to process long sequences of tokens such as in videos \cite{swiftformer, khan2024crop}. To alleviate this issue, \cite{swiftformer} introduced a redesigned attention mechanism named Efficient Additive Attention as seen in \cref{fig:7} (a). This proposed mechanism replaces the expensive matrix multiplication operations with element-wise multiplications and linear transformations with the use of only the key-value interaction. However, such methods have not yet been investigated for the task of violence detection to the best of our knowledge. This poses an opportunity to modify and enhance the concepts discussed to create an improved, tailor-made solution for the problem of violence detection. 

\subsection{CUE-Net Architecture}

We introduce our novel architecture, the spatial \textbf{C}ropping, enhanced \textbf{U}niformerV2 with Modified \textbf{E}fficient Additive Attention network (\textbf{CUE-Net}) for violence detection in videos as shown in \cref{fig:3}.  The architecture contains five main components, namely: (a) Spatial Cropping Module; (b) 3D Convolution Backbone; (c) Local UniBlock V2; (d) Global UniBlock V3; and (e) Fusion Block, inspired by the motivational factors discussed in the preceding paragraph.










\begin{algorithm}
\caption{Spatial Cropping Mechanism with YOLO V8}\label{alg:cap}
\begin{algorithmic}
\Require $x$ \Comment{Input Video}
\Ensure $x'$ \Comment{Cropped Video}
\\ \Comment{A crop box for a video is defined by the coordinates $(x_{min}, y_{min})$ and $(x_{max}, y_{max})$ }
\State $(x_{min}, y_{min}) \gets (inf, inf)$ 
\State $(x_{max}, y_{max}) \gets (0, 0)$ 
\State $max\_people \gets 0$  \Comment{Max no of people}
\\

\State $F \gets YOLO\_V8(x) $ 
\Comment{$F$ is the list of frames of the video where the $i$th entry $f_i$ is another list of people bounding boxes $P$ found in each frame, the $j$th bounding box $p_j$ denoted as $(x^j_{min}, y^j_{min}), (x^j_{max}, y^j_{max})$}.

\\
\For{each $f_i$ in $F$}
    \State $n_i \gets 0$  \Comment{no. of people in each frame}
    \For{each $p_j$ in $f_i$}
        \State $x_{min} \gets \min( x_{min}, x^j_{min})$
        \State $y_{min} \gets \min(y_{min},  y^j_{min})$
        \State $x_{max} \gets \min( x_{max}, x^j_{max})$
        \State $y_{max} \gets \min(y_{max},  y^j_{max})$
        \State $people \gets people+1$
    \EndFor

    \If{$n_i > 0$}
        \State $max\_people \gets max(max\_people, n_i)$
    \EndIf
\EndFor

\If{$max\_people > 1$}
    \State $x' \gets crop\_video(x_{min}, y_{min}, x_{max}, y_{max})$
\Else
    \State $x' \gets x$
\EndIf

\end{algorithmic}
\end{algorithm}

\subsubsection{Spatial Cropping Module}

The motivation for cropping the video spatially is based on the observation that violence is normally carried out between two or more people. We opted to extract the people and crop the video frames spatially with the maximum bounding box for the area where people are found, so as not to lose the information surrounding the people, but to maximize the important area to focus by removing the parts of the environment where the people are not present. We opted not to perform temporal cropping  to avoid any information loss occurring from undetected people. When the video $\mathbf{X}\in \R^{T\times H\times W\times c}$ ($T$, $H$, $W$ and $c$ represent temporal dimensions, height, width and colour channels of the video frames respectively) is input to this spatial cropping module, to detect people, we used the YOLO (You Only Look Once) V8 algorithm \cite{yolov8} which classifies objects in a single pass using a CNN-based architecture where a full image is taken as the input. \cref{alg:cap} elaborates the spatial cropping procedure for the maximum bounding box throughout the video. If more than one person is detected, it outputs $\mathbf{X'}\in \R^{T\times H\times W\times c}$ which is the spatially cropped video. If only a single person or no people are detected, $\mathbf{X'}$ will be the initial video as a whole to make sure the method does not miss out any information.

\subsubsection{3D Convolution Backbone}

The spatially cropped video frames from the previous module $\mathbf{X'}$ are then passed as input to the 3D Convolution Backbone, where a 3D convolution (i.e., 3$\times$16$\times$16) is used to encode and project the input video as spatio-temporal tokens $\mathbf{V}^{0}\in \R^{T\times H\times W\times d}$, ($T$, $H$, $W$ and $d$ represent temporal dimensions, height and width of the frames and hidden dimensions respectively). Afterwards, according to the original ViT design \cite{vit_main}, spatial downsampling by $16\times$ is performed and then a temporal downsampling by $2\times$ is performed to reduce spatio-temporal resolution. The encoded hidden dimension $d$ was maintained the same throughout the architecture modules to facilitate residual connections. At the end of this stage the processed input is sent to the Local UniBlock V2.

\subsubsection{Local UniBlock V2}

The Local UniBlock V2, has been introduced specifically to model the local dependencies in our CUE-Net architecture. This was extracted from the UniformerV2 architecture without modifications as a result of the ablation study we performed. Here, two types of Multi-Head Relation Aggregator (MHRA) units are used namely, Local Temporal MHRA (LT\_MHRA) and Global Spatial (GS\_MHRA) along with a Feed Forward Network (FFN) module. The input to this block is $\mathbf{V}^{0}\in \R^{T\times H\times W\times d}$ which is the output of the previous 3D Convolution Backbone and this block outputs $\mathbf{V}^{3}\in \R^{T\times H\times W\times d}$ at the end of the FFN. The processing inside a Local UniBlock V2 can be represented as:
\begin{align}
\begin{split}
    \mathbf{V}^{1} ={}&
        \mathbf{V}^{0} +
        {\rm LT\_MHRA}
        \left(
            {\rm LN}
            \left(
                \mathbf{V}^{0}
            \right)
        \right) 
        , 
\end{split}\\
\begin{split}
    \mathbf{V}^{2} ={}&
         \mathbf{V}^{1} +
        {\rm GS\_MHRA}
        \left(
            {\rm LN}
            \left(
                \mathbf{V}^{1}
            \right)
        \right) 
        ,
\end{split}\\ 
   \mathbf{V}^{3} ={}&
        \mathbf{V}^{2} +
        {\rm FFN}
        \left(
            {\rm LN}
            \left(
                \mathbf{V}^{2}
            \right)
        \right) 
        , 
\end{align}
where ${\rm LN}(\cdot)$ represents layer normalization. A Multi-Head Relation Aggregator (MHRA) unit concatenates multiple heads and can be described as:
\begin{align}
\begin{split}
     {\rm S}_n(\mathbf{V^i}) ={}&
        {\rm \mathbf{B}}_n\cdot{\rm L}_n(\mathbf{V^i}), \label{a}
\end{split}\\
    {\rm MHRA}(\mathbf{V^i}) ={}&
        [
            {\rm S}_1(\mathbf{V^i}); 
            {\rm S}_2(\mathbf{V^i}); 
            \cdots; 
            {\rm S}_N(\mathbf{V^i})
        ]
        \cdot\mathbf{M},
\end{align}
where the relational aggregator of the $n$-th head is represented by ${\rm S}_n(\cdot)$ where $\mathbf{B}_n$ 
represents an affinity matrix that characterizes the relationships between tokens and $\mathbf{B}_n$ is changed accordingly in LT\_MHRA and in GB\_MHRA to achieve their respective goals. A linear projection is represented by ${\rm L}_n(\mathbf{\cdot})$. A fusion matrix $\mathbf{M}\in \R^{d\times d}$ which is learnable  is used to integrate $N$ heads during concatenation of the heads represented by $[...]$ at the end of a general MHRA unit.

\textbf{LT\_MHRA}: The Local Temporal MHRA (LT\_MHRA) takes the input $\mathbf{V}^{0}$ from the 3D Convolution Backbone, implements depth-wise convolution (DWConv) with the help of the affinity matrix $\mathbf{B}_n$ described in the preceding paragraph, as the goal of this unit is to reduce the local temporal redundancy and to learn local representations form the local spatio-temporal context. This unit outputs $\mathbf{V}^{1}\in \R^{T\times H\times W\times d}$.

\textbf{GT\_MHRA}: The Global Temporal MHRA (GT\_MHRA) receives the output of the LT\_MHRA unit $\mathbf{V}^{1}$ and implements multi-headed self-attention (MHSA) from the ViT architecture \cite{vit_main} with the help of the affinity matrix $\mathbf{B}_n$ described earlier as the goal of this unit is to make use of the rich image pretraining of ViTs learned from large image databases. To achieve this target, the GT\_MHRA units are initialized with image-pretrained ViT embeddings inflated along the temporal dimension and the output of this unit is $\mathbf{V}^{2}\in \R^{T\times H\times W\times d}$.

\textbf{FFN}: The Feed Forward Network (FFN) module accepts the output $\mathbf{V}^{2}$ of GT\_MHRA, and consists of two linear projections separated by a GeLU \citep{gelu} activation function.  FFN is implemented at the end of the Local UniBlock V2 to output $\mathbf{V}^{3}\in \R^{T\times H\times W\times d}$.

\subsubsection{Global UniBlock V3}

The Global UniBlock V3 has been introduced specifically to perform global long-range dependency modeling on the spatio-temporal scale in our CUE-Net. This Global UniBlock V3 consists of three basic units namely, Dynamic Positional Embedding (DPE) unit, Modified Efficient Additive Attention (MEAA) unit, and finally a Feed Forward Network (FFN) module. The input to this block is $\mathbf{V}^{3}\in \R^{T\times H\times W\times d}$ which is the output of the previous Local UniBlock V2 and the Global UniBlock V3 outputs $\mathbf{V}^{6}\in \R^{1\times d}$ at the end of the FFN unit. The processing inside this block where ${\rm LN}(\cdot)$ represents layer normalization can be represented as:
\begin{align}
\begin{split}
    \mathbf{V}^{4} ={}&
         \mathbf{V}^{3} +
        {\rm DPE}
        \left(
            \mathbf{V}^{3}
        \right) 
        ,
\end{split}\\
\begin{split}
    \mathbf{V}^{5} ={}&
        {\rm MEAA}
        \left(
            {\rm LN}
            \left(
                \mathbf{q}
            \right), 
            {\rm LN}
            \left(
                \mathbf{V}^{4}
            \right)
        \right), 
\end{split}\\ 
   \mathbf{V}^{6} ={}&
         \mathbf{V}^{5} +
        {\rm FFN}
        \left(
            {\rm Norm}
            \left(
                \mathbf{V}^{5}
            \right)
        \right) 
        .  
\end{align}
\textbf{DPE}: The Dynamic Positional Embedding (DPE) unit receives the input $\mathbf{V}^{3}$ from the previous Local UniBlock V2, and uses simple 3D depth-wise spatio-temporal convolution with zero padding (DWConv) to encode spatio-temporal positional information for token representations, as the videos vary both spatially and temporally. The output of the DPE block is $\mathbf{V}^{4}\in \R^{T\times H\times W\times d}$.

\begin{figure}[t]
  \centering
   \includegraphics[width=1\linewidth]{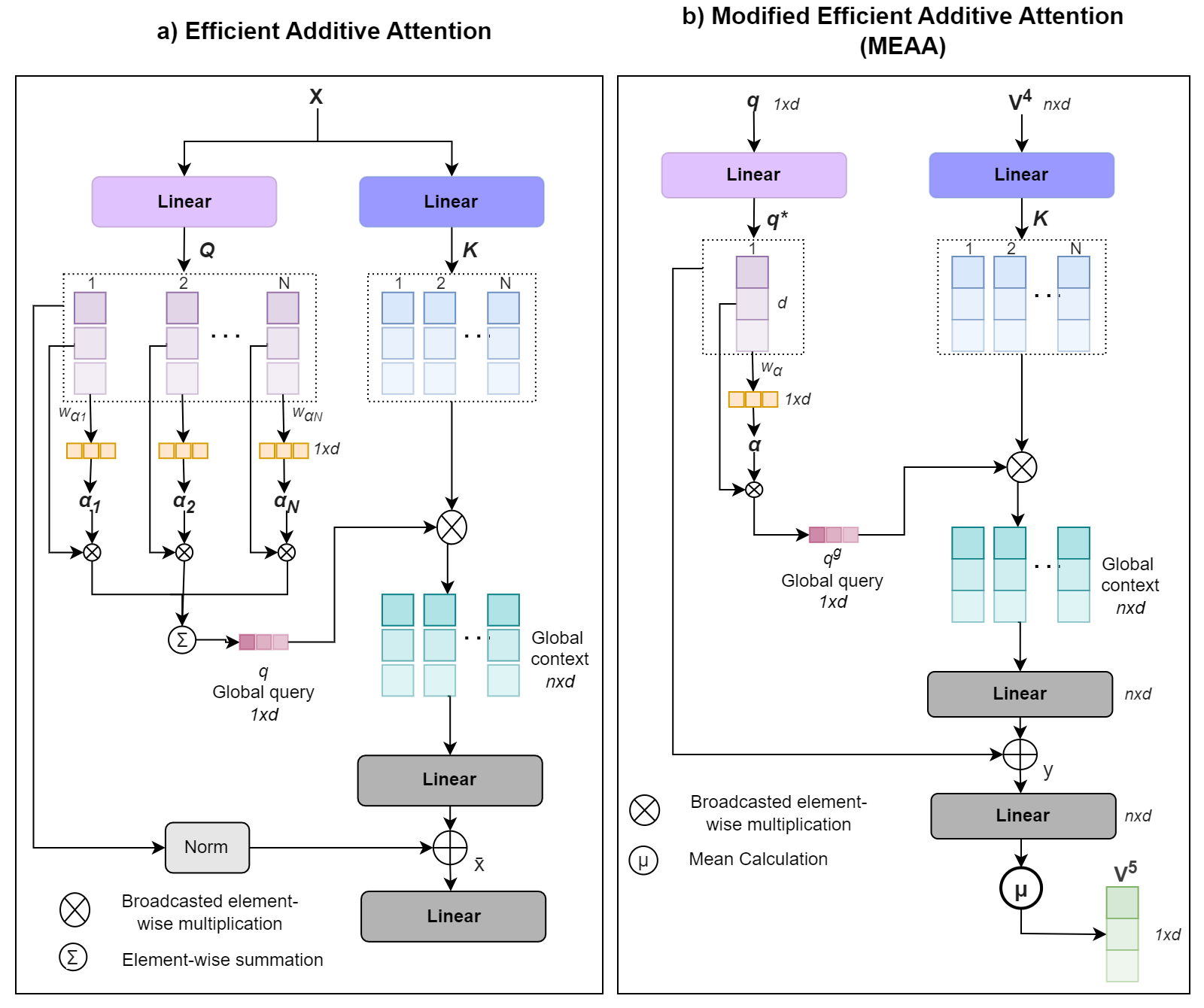}

   \caption{\textbf{(a)} illustrates the Efficient Additive Attention where the expensive matrix multiplication operations have been replaced with element-wise multiplications and linear transformations via a query-key pair interaction. \textbf{(b)} represents the Modified Efficient Additive Attention (MEAA) which only uses a query vector instead of a whole query matrix when computing Efficient Additive Attention, reducing the computational complexity along with memory usage. }
   \label{fig:4}
\end{figure}

\textbf{Modified Efficient Additive Attention (MEAA)}: In the Modified Efficient Additive Attention (MEAA) unit, a learnable query $\mathbf{q} \in \R^{1\times d}$ is converted into a video representation,
through modeling a relationship between this query $\mathbf{q}$ and all the spatio-temporal tokens $\mathbf{V^4}$ received from the DPE unit, with the help of this modified version of Efficient Additive Attention. As depicted in \cref{fig:4} (b), the learnable query vector $ \mathbf{q}$ is projected into query ($\mathbf{q}^{*}$) and $\mathbf{V}^{4}$ is projected into the key ($\mathbf{K}$) using two linear layers where $n$ is the token length and $d$ is the number of hidden dimensions. Afterwards, another vector of learnable parameters $\mathbf{w}_a$ $\in \mathbb{R}^{d}$ is multiplied with the query $\mathbf{q}^{*}$ with the intention of learning the attention weights of the query. This results in outputting  $\alpha$ $\in \mathbb{R}^{1}$ which can be considered the global attention query vector:
\begin{equation}\label{eq1}
\alpha  =  \Bigl(\frac{\mathbf{q}^{*} \cdot \textbf{w}_{a}}{\sqrt{d}}\Bigl)
\end{equation}
The global query vector $\mathbf{q}^{g}$ $\in \mathbb{R}^{1 \times d}$ is afterwards derived using the attention weight which was learned as:
 \begin{equation}
    \mathbf{q}^{g}=\ \alpha \odot \mathbf{q}^{*},
\end{equation}
where $\odot$ represents element-wise multiplication.

Finally, element-wise multiplication is performed between the global query vector $\mathbf{q}^{g}$ $\in \mathbb{R}^{1 \times d}$  and the key matrix $\mathbf{K}$ $\in \mathbb{R}^{n\times d}$ in order to fuse these two entities, where the end result has dimensions $\mathbb{R}^{n\times d}$. The above process is inexpensive, with linear complexity in relation to token length, compared to obtaining self-attention which has a quadratic complexity. A linear layer is then applied to this element-wise multiplication with a residual connection from $\mathbf{q}^{*}$ along with a final linear layer to produce the output:
\begin{equation}
  {\mathbf{V}^{5}}= \mathbf{Mean}\big(\mathbf{W}_{2} \cdot((\mathbf{W}_1\cdot( \mathbf{q}^{g} \odot \mathbf{K})+\mathbf{b}_{1}) + \mathbf{q}^{*})+\mathbf{b}_{2}\big). 
\end{equation}
To obtain $\mathbf{V}^{5} \in \R^{1\times d}$ as the output, the mean is calculated along the $n$ dimension to get an overall representation. 

\textbf{FFN}: Similar to the FFN module in the previous Local UniBlock v2, this Feed Forward Network (FFN) accepts the output $\mathbf{V}^{5}$ of GT\_MHRA module and consists of two linear projections separated by a GeLU \citep{gelu} activation function, at the end of the Global UniBlock V3 to output $\mathbf{V}^{6}\in \R^{1\times d}$.

\subsubsection{Fusion Block} 

At the very end of the CUE-Net architecture, a fusion block integrates the final token from the Global UniBlock $\mathbf{V}^{6}\in \R^{1\times d}$ with the final video class token $\mathbf{V}^{3'}\in \R^{1\times d}$ extracted from the final output $\mathbf{V}^{3}\in \R^{T\times H\times W\times d}$ of the Local UniBlock. These tokens $\mathbf{V}^{6}$ and $\mathbf{V}^{3'}$ are dynamically fused to obtain $\textbf{Z}$ as:
\begin{align}
\begin{split}
    \beta' ={}&
         Sigmoid(\beta)
        ,
\end{split}\\
   \mathbf{Z} ={}&
         (1-\beta')\odot\mathbf{V}^{6} +
        \beta'\odot\mathbf{V}^{3'} 
        .  
\end{align}
using another learnable parameter $\beta \in \R^{1\times d}$ passed through the Sigmoid function. Finally, the target class $Pr$ is obtained by passing $\textbf{Z}$ through a fully connected projection layer.

\section{Experiments and Results}
\label{sec:exp}

\subsection{Datasets}

The most challenging datasets so far in the VD domain are the Real-World Fighting (RWF-2000) dataset \cite{rwf_dataset} and the Real Life Violence Situations (RLVS) dataset \cite{reallife_violence_dataset}, that contain video footage of fighting in real life scenarios. But of these two datasets, only the RWF-2000 dataset contains exclusive surveillance footage.

\subsubsection{Real World Fighting (RWF-2000) Dataset}

The Real World Fighting (RWF-2000) dataset \cite{rwf_dataset} was introduced in 2020 and is the most comprehensive dataset, containing real world fighting scenarios sourced purely through surveillance footage.  A typical violent example can be seen at \cref{fig:7} (b). RWF-2000 contains 2,000 trimmed video clips captured by surveillance cameras from real-world scenes collected from YouTube. Each video is trimmed to 5 seconds where the fighting occurs. The dataset is  balanced with 1000 violent videos and 1000 non-violent videos, with a 80\%-20\% predefined train-test split which has been thoroughly checked for data leakage between the splits.

\subsubsection{Real Life Violence Situations (RLVS) Dataset}

The Real Life Violence Situations (RLVS) dataset \cite{reallife_violence_dataset} consists of 2000 video clips with 1000 violent and another 1000 non-violent videos collected from YouTube. These contain many real street fight situations in several environments and conditions with an average length of 5s from different sources such as surveillance cameras, movies, video recordings, etc. Similar to RWF-2000, a 80\%-20\% train-test split has been created for this dataset.

\begin{table}
  \centering
  \begin{tabular}{ p{2.25cm} p{2.5cm} p{2cm}}
    \toprule
    \makecell*[c]{Method} & \makecell*[c]{Model Type} & \makecell*[c]{Accuracy (\%)}\\
    \midrule
    \makecell*[c]{ConvLSTM\cite{rwf_dataset}} & \makecell*[c]{CNN+LSTM} & \makecell*[c]{77.00} \\
    \makecell*[c]{X3D\cite{swintransformer_VD}} & \makecell*[c]{3DCNN} & \makecell*[c]{84.75} \\
    \makecell*[c]{I3D\cite{action_recognition_VD_5}} & \makecell*[c]{3DCNN} & \makecell*[c]{83.40} \\
    \makecell*[c]{Flow Gated \\ Network\cite{rwf_dataset}} & \makecell*[c]{Two Stream \\ Graph CNN} & \makecell*[c]{87.25} \\
    \makecell*[c]{SPIL\cite{action_recognition_VD_3}} & \makecell*[c]{Graph CNN} & \makecell*[c]{89.30} \\
   \makecell*[c]{Structured \\ Keypoint \\ Pooling\cite{action_recognition_VD_5}} & \makecell*[c]{CNN} & \makecell*[c]{93.40} \\
   \makecell*[c]{Video Swin \\ Transfor-\\mer\cite{videoswintrf}} & \makecell*[c]{ViViT} & \makecell*[c]{91.25} \\
   \makecell*[c]{ACTION-\\VST\cite{swintransformer_VD}} & \makecell*[c]{CNN + ViViT} & \makecell*[c]{93.59}\\

    \makecell*[c]{\textbf{CUE-Net} \\ \textbf{(Ours)}} & \makecell*[c]{\textbf{Enhanced} \\\textbf{UniformerV2} \\\textbf{+ MEAA}}  & \makecell*[c]{\textbf{94.00}}\\
    \bottomrule
  \end{tabular}
  \caption{Results comparison for the RWF-2000 Dataset.}
  \label{tab:rwf}
\end{table}

\subsection{Implementation Details}

Our algorithm was implemented in PyTorch using the AdamW optimizer \cite{adamw} with a cosine learning rate schedule \cite{cosinelnr} starting with a learning rate of 1e-5 and Cross-Entropy Loss, taking insights from training recipes of the original UniformerV2 architecture \cite{uniformerv2}. To initialize the Global MHRA units of the Local UniBlocks, pretrained embeddings from CLIP-ViT \cite{clip_model} model are used as \cite{uniformerv2} states this yields the best results in their architecture due to the well learned representations by vision-language contrastive learning. All models were trained for 50 epochs where the best validation model was saved after each epoch.  We utilized NVidia A100 GPUs with 40GB/80GB memory. For data augmentation, RandAugment by \cite{randaugment} was used. Our best performing CUE-Net architecture consisted of 354M parameters where the number of frames selected ($T$) to be inputted was 64 with a resized frame height ($H$) and width ($W$) of 336 $\times$ 336 in RGB channels ($c=3$). 

\subsection{Results}

In this section we perform an in-depth analysis comparing our CUE-Net architecture with other leading architectures using the two different datasets, RWF-2000 and RLVS. Following the practice of other researchers \cite{action_recognition_VD_5, swintransformer_VD}, we also use classification accuracy as the metric to evaluate the performance as both of the trained and tested upon datasets are balanced. \cref{tab:rwf} and \cref{tab:rlvs} present the results comparison of our CUE-Net architecture with other state-of-the-art methods on RWF-2000 and RLVS datasets respectively. Our CUE-Net architecture outperforms all others in classification accuracy.  On the RWF-2000 dataset, our CUE-Net architecture reaches an accuracy of 94.00\%, and on the RLVS dataset, it records an accuracy of 99.50\%, setting a new state-of-the-art on both datasets.

\begin{table}
  \centering
  \begin{tabular}{ p{2.25cm} p{2.5cm} p{2cm}}
    \toprule
    \makecell*[c]{Method} & \makecell*[c]{Model Type} & \makecell*[c]{Accuracy (\%)}\\
    \midrule
    \makecell*[c]{CNN-\\LSTM\cite{reallife_violence_dataset}} & \makecell*[c]{VGG16+LSTM} & \makecell*[c]{88.20} \\
    
    \makecell*[c]{Temporal \\ Fusion CNN \\+LSTM\cite{tempfusion}} & \makecell*[c]{CNN+LSTM} & \makecell*[c]{91.02} \\
    
    \makecell*[c]{DeVTr\cite{devtr}} & \makecell*[c]{ViViT} & \makecell*[c]{96.25} \\

    \makecell*[c]{ACTION-\\VST\cite{swintransformer_VD}} & \makecell*[c]{CNN + ViViT} & \makecell*[c]{98.69}\\

    \makecell*[c]{\textbf{CUE-Net} \\ \textbf{(Ours)}} & \makecell*[c]{\textbf{Enhanced} \\\textbf{UniformerV2} \\\textbf{+ MEAA}}  & \makecell*[c]{\textbf{99.50}}\\
   
    \bottomrule
  \end{tabular}
  \caption{Results comparison for the RLVS Dataset.}
  \label{tab:rlvs}
\end{table}

\subsubsection{Visual Analysis of Results}

\textbf{RWF-2000 Dataset}: For the RWF-2000 test set, we performed a visual evaluation of the misclassified instances. As the accuracy was 94.00\%, there were only 24 misclassified instances where 15 non-violent videos were misclassified as violent and 8 violent videos were misclassified as non-violent. This gives the idea that our model is better able to learn the specifics of the violent action markers. Supporting this proposition, we were able to identify a video shown in \cref{fig:7} (c) where a man makes punching actions but is not really engaging in a fight. Our method misclassifies this non-violent video as a violent video.

\textbf{RLVS Dataset}: We also performed a visual evaluation of the misclassified instances in RLVS test set. Since our accuracy was 99.5\%, there were only 2 misclassified videos where 1 non-violent video was misclassified and vice versa. When analysing the 2 misclassified videos, we noted the video shown in \cref{fig:7} (d) was labelled violent and was misclassified, but was a \emph{mislabeled instance of a non-violent video} where two players were playing tennis without any violence, thus increasing the true accuracy of our model with this correction to 99.75\%. This strongly shows CUE-Net has learned the dynamics of violent actions.

\subsection{Ablation Study}

We performed a series of ablation studies to asses the efficacy of the components of CUE-Net.  

\subsubsection{Ablation on Spatial Cropping, Self-Attention and MEAA in Local UniBlock and Global UniBlock}

\begin{table}
  \centering
  \resizebox{0.48\textwidth}{!}{
  \begin{tabular}{p{1.33cm} p{1.45cm} p{1.45cm} p{1.4cm} p{1.12cm}}
    \toprule
    \makecell[c]{Spatial \\Cropping} & \makecell[c]{Local \\UniBlock} & \makecell[c]{Global \\UniBlock} & \makecell[c]{Accuracy \\(\%)} & \makecell[c]{FLOPs \\(Giga)}\\
    \midrule
    
    \makecell[c]{$\times$} & \makecell[c]{Self- \\Attention} & \makecell[c]{Self- \\Attention} &  \makecell*[c]{92.00} & \makecell[c]{6108}\\

    \makecell[c]{$\checkmark$} & \makecell[c]{Self- \\Attention} & \makecell[c]{Self- \\Attention} &  \makecell*[c]{92.50} & \makecell[c]{6108} \\

    \makecell[c]{$\checkmark$} & \makecell[c]{MEAA} & \makecell[c]{Self- \\Attention} &  \makecell*[c]{50.00} & \makecell[c]{5929} \\

    \makecell[c]{$\checkmark$} & \makecell[c]{MEAA} & \makecell[c]{MEAA} &  \makecell*[c]{50.00} & \makecell[c]{\textbf{5749}} \\

    \makecell[c]{\textbf{$\checkmark$}} & \makecell[c]{\textbf{Self-} \\\textbf{Attention}} & \makecell[c]{\textbf{MEAA}} &  \makecell*[c]{\textbf{94.00}} & \makecell[c]{5826} \\

    \bottomrule
  \end{tabular}
  }
  \caption{Ablation showing the effect of Spatial Cropping, Self-Attention and MEAA in the Local UniBlock and Global UniBlock.}
  \label{tab:abl1}
\end{table}

Four ablation experiments were conducted to explore the use of spatial cropping and the MEAA module as shown in in \cref{tab:abl1}. First, we remove the spatial cropping module, and use Self-Attention both in the Local UniBlock and in the Global UniBlock. In the second row of the table, we add spatial cropping, which enhances the performance of the model.  In the last row, we replace the Self-Attention with Modified Efficient Additive Attention (MEAA) in the Global UniBlock, forming our full CUE-Net model. This provides a considerable boost of 1.5\% in accuracy.  We speculate traditional Self-Attention may have an information overload especially while trying to capture representative features temporally. In contrast, with the simpler MEAA, it may be easier for the Global UniBlock to learn the discriminative features temporally when it comes to identifying violent actions. The remaining rows in the table explore the use of MEAA in the Local UniBlock.  Here the algorithm performance becomes random as shown by the results in \cref{tab:abl1}. In this setting, the local UniBlocks are not initialized with pretrained ViT embeddings and underperform.  Also, it is evident from \cref{tab:abl1} that the FLOPs count reduces when MEAA is used in place of Self Attention depicting a reduction in computational complexity. Therefore our proposed approach of using Self-Attention in the Local UniBlock and MEAA in the Global UniBlock has the best performance along with a reduced FLOPs count.


\subsubsection{Ablation on Original Efficient Additive Attention vs Modified Efficient Additive Attention (MEAA)}

We also experimented with the original Efficient Additive Attention with a n-dimensional query matrix instead of a 1-dimensional query vector in the Global UniBlock in our CUE-Net architecture, but it under-performed, with 1\% less accuracy compared to MEAA as seen in \cref{tab:abl2}. We also note the GPU memory consumption was considerably higher (47 GB compared to 35 GB) when the original Efficient Additive Attention was used. Therefore, we can state that our MEAA gives a competitive edge over original Efficient Additive Attention when it comes to memory usage.

\begin{table}
  \centering
  \begin{tabular}{ p{3.5cm} p{1.5cm} p{1.75cm}}
    \toprule
    \makecell[c]{ Efficient Additive \\Attention Variant} & \makecell[c]{Accuracy \\(\%)} & \makecell[c]{GPU \\Memory\\Usage}\\
    \midrule
    
    \makecell[c]{Original} & \makecell*[c]{93.00} & \makecell*[c]{47.33 GB}\\

    \makecell[c]{\textbf{MEAA}} & \makecell*[c]{\textbf{94.00}} & \makecell*[c]{\textbf{35.04 GB}} \\

    \bottomrule
  \end{tabular}
  \caption{Ablation showing the Original Efficient Additive Attention vs Modified Efficient Additive Attention (MEAA).}
  \label{tab:abl2}
\end{table}
\section{Conclusion}
\label{sec:conclusion}

This paper introduces CUE-Net, a novel framework for violence detection in videos which implements cropping with an enhanced version of UniformerV2 architecture. CUE-Net uses convolution-based mechanisms to capture the local features and attention mechanisms to capture the global spatio-temporal features fused with a novel attention mechanism named Modified Efficient Additive Attention. We incorporated video cropping spatially, based on the detected number of people before the video is fed into the main processing algorithm to focus the method on the areas where violence takes place. We also proposed Modified Efficient Additive Attention instead of Self-Attention in the Global UniBlock V3 of the CUE-Net architecture, to capture the important global spatio-temporal features, as it has shown to be effective and efficient. Our proposed CUE-Net algorithm has achieved new state-of-the-art performance on the RWF-2000 and RLVS datasets, surpassing the results of most recently published methods. 

\textbf{Acknowledgment.} This work has been funded through an EPSRC DTP studentship at Queen Mary University of London in collaboration with Remark AI UK Ltd. This paper utilised Queen Mary's Andrena HPC facility, supported by QMUL Research-IT Services \cite{apocrita}.
{
    \small
    \bibliographystyle{ieeenat_fullname}
    \bibliography{main}
}


\end{document}